\documentclass{article}

\usepackage{PRIMEarxiv}

\usepackage[utf8]{inputenc} %
\usepackage[T1]{fontenc}    %
\usepackage{hyperref}       %
\usepackage{url}            %
\usepackage{booktabs}       %
\usepackage{amsfonts}       %
\usepackage{nicefrac}       %
\usepackage{microtype}      %
\usepackage{lipsum}
\usepackage{fancyhdr}       %
\usepackage{graphicx}       %
\graphicspath{{media/}}     %

\usepackage[utf8]{inputenc} %
\usepackage{algorithm2e}
\usepackage{array}
\usepackage{algpseudocode}
\usepackage{textcomp}
\usepackage{verbatim}
\usepackage[T1]{fontenc}    %
\usepackage{url}            %
\usepackage{booktabs}       %
\usepackage{amsfonts}       %
\usepackage{nicefrac}       %
\usepackage{microtype}      %
\usepackage{xcolor}         %
\usepackage{soul}

\usepackage{amsmath}
\usepackage{amssymb}
\usepackage{amsthm}
\usepackage{bm}
\usepackage{color}
\usepackage{graphicx}
\usepackage{makecell}
\usepackage{multirow}
\usepackage{tabularx}
\usepackage{mathtools}
\usepackage{pythonhighlight}
\usepackage{listings}

\definecolor{pink}{rgb}{0.858, 0.188, 0.478}
\usepackage{xspace}

\definecolor{commentcolor}{RGB}{110,154,155}   %

\usepackage{natbib}
  
\title{DeInfoReg: A Decoupled Learning Framework with Information Regularization for Better Training Throughput
}

\author{
  Zih-Hao Huang, You-Teng Lin, Hung-Hsuan Chen \\
  Computer Science and Information Engineering \\
  National Central University \\
  Taoyuan, Taiwan\\
  \texttt{ianzih0730@gmail.com, lyt0310603@gmail.com, hhchen1105@acm.org}
}

\begin{document}
\maketitle

\begin{abstract}

This paper introduces Decoupled Supervised Learning with Information Regularization (DeInfoReg), a novel approach that transforms a long gradient flow into multiple shorter ones, thereby mitigating the vanishing gradient problem. Integrating a pipeline strategy, DeInfoReg enables model parallelization across multiple GPUs, significantly improving training throughput. We compare our proposed method with standard backpropagation and other gradient flow decomposition techniques. Extensive experiments on diverse tasks and datasets demonstrate that DeInfoReg achieves superior performance and better noise resistance than traditional BP models and efficiently utilizes parallel computing resources. The code for reproducibility is available at: \url{https://github.com/ianzih/Decoupled-Supervised-Learning-for-Information-Regularization/}.

\end{abstract}

\keywords{
backpropagation, local loss, contrastive learning}

\section{Introduction} \label{sec:intro}

Deep neural networks have trended towards increasing layers to enhance the recognition capabilities of models. However, increasing the number of layers lengthens the flow of the gradient, leading to vanishing gradients~\citep{Hochreiter1998VanishingGradient}. Although some solutions have been proposed, such as the residual modules in ResNet~\citep{he2016deep} and the inception modules in GoogleNet~\citep{szegedy2015going}, these methods only partially address these problems because the long gradient flows still exist in the network. The long gradient flow also causes gradient locking~\citep{jaderberg2017decoupled}, which prevents layers from updating independently and hinders training performance since the whole forward-backward learning approach requires $O(L)$ for each batch ($L$ is the number of layers).

Meanwhile, a new model paradigm, which we call the Decoupled Paradigm, decouples a deep neural network into multiple learning modules~\citep{nokland2019predsim, lee2015deeply, sid2023blockwise}; each module has a local loss function, and the gradient flows between different modules are cut off. As a result, gradient flows are short even if a network is deep, so problems of vanishing gradients are less likely to occur~\citep{kao2021associated, wu2021associated}. For example, Associated Learning (AL) converts features $x$ and targets $y$ into similar latent representations as a block, then recursively transforms these into meta-representations as new blocks, with a local objective function in each block. In another model, Supervised Contrastive Parallel Learning (SCPL)~\citep{wang21scpl}, the contrastive loss is the local loss for parameter updates. Additionally, when incorporated with the pipeline strategy, training performance can be improved as the parameters in different layers (allocated to different computing units) can be updated simultaneously.

Although the models based on the Decoupled Paradigm have the above advantages~\citep{kao2021associated, wu2021associated, wang21scpl, ho22scpl}, the decoupled model brings new challenges. First, decoupled models usually lead to less accurate predictions than standard backpropagation (BP). Second, 
many decoupled models are based on contrastive learning, which still faces challenges such as the collapse problem and the need for large negative sample pairs~\citep{khosla2020SupCon, ozsoy2022corinfomax}.
Third, while decoupled models have the potential to leverage pipelining for model parallelism, practical implementations have not been officially released due to non-trivial engineering challenges.

This paper makes the following contributions to address the above challenges.

\begin{itemize}
    \item We introduce DeInfoReg, a novel decoupled model architecture that leverages self-supervised learning (SSL) as the backbone. Our experiments are conducted on open datasets for natural language processing and computer vision. Our empirical results demonstrate that DeInfoReg achieves test accuracies on open datasets that are comparable to -- or in some cases even exceed -- those obtained with traditional backpropagation (addressing the first challenge). 

    \item We incorporate information regularization techniques within the loss function to mitigate the collapse problem. Experimental results demonstrate that a large batch size is not required (addressing the second challenge).

    \item We observed a larger magnitude of gradient values than BP during training, demonstrating improved stability in deep structures and the ability to mitigate issues such as gradient vanishing (explaining why DeInfoReg may address the first challenge).

    \item DeInfoReg tolerates noisy labels better than BP. This empirical discovery indicates the potential of DeInfoReg to handle complex tasks with greater efficiency and reliability (more empirical evidence on the superiority of DeInfoReg for challenge 1). 

    \item We integrate DeInfoReg with pipelining and demonstrate that this integration enables model parallelism. Thus, our approach significantly improves training efficiency and scalability, facilitating the deployment of more complex models in real-world applications (addressing challenge 3).
\end{itemize}

The rest of the paper is organized as follows. Section~\ref{sec:rel-work} reviews previous studies on gradient flow decoupling and contrastive learning, which is the foundation of DeInfoReg. Section~\ref{sec:method} explains our proposed DeInfoReg model in detail. The experimental results are reported in Section~\ref{sec:exp}. Finally, we conclude and discuss the work in Section~\ref{sec:disc}. 

\section{Related Work} \label{sec:rel-work}

This section reviews previous work on decoupling gradient flows. We also review studies of contrastive learning and model parallelism.

\subsection{Decoupled Paradigm}

The decoupled paradigm decomposes a deep neural network into multiple modules, each with its own local loss function, effectively blocking the gradient flow between modules.

Previous research using local losses or truncated gradient flows to modularize neural networks has been motivated by various factors, such as reducing memory usage~\citep{wang2021revisiting, yang2024towards} and allowing faster inference~\citep{teerapittayanon2016branchynet}. However, most of these approaches have been tested only on relatively simple network architectures, and their performance has generally lagged behind that of standard backpropagation. Only in limited decoupled models, such as Successive Gradient Reconciliation (SGR)~\citep{yang2024towards}, AL~\citep{kao2021associated, wu2021associated}, and SCPL~\citep{wang21scpl, ho22scpl}, show performance comparable to BP. Moreover, to our knowledge, no publication has yet combined decoupled models with pipelining to achieve model parallelism for enhanced training throughput.

\subsection{Contrastive Learning}

Contrastive learning (CL) has become an important research focus due to its ability to leverage large amounts of unlabeled data to learn object representations~\citep{khosla2020SupCon, ozsoy2022corinfomax, oord2018INFONCE, chen2021simsiam, he2020moco, chen2020mocov2, chen2020simclr, zbontar2021barlow}. However, CL usually suffers from the collapse problem~\citep{ozsoy2022corinfomax, jing2021collapse}: different input features are mapped to the same output vector space, resulting in a loss of distinction. Methods to mitigate this issue often require sacrificing computational efficiency or increasing the complexity of the model, such as using large batch sizes~\citep{chen2020simclr} or memory banks~\citep{he2020moco} to maintain diverse representations.

Contrastive learning can be effectively combined with multi-modal learning, where models process inputs such as images with accompanying text or audio. This integration enables the development of robust and semantically rich representations that capture complementary information across modalities. Such a fusion not only enhances the model's capacity to distinguish similar concepts but also improves its generalization across tasks involving varied data sources. Multimodal contrastive learning frameworks -- illustrated by works like CLIP~\citep{radford2021learning} and ALIGN~\citep{jia2021scaling} -- achieve cross-modal consistency by aligning representations of different modalities/forms in a shared latent space. These modalities, in a broader sense, can include data of different formats (e.g., text, images, or videos), spatial and temporal representation coherence~\citep{shu2021spatiotemporal, tang2019coherence}, or different granularity representations of an object~\citep{shu2022multi, yan2020higcin}. Furthermore, by leveraging the inherent diversity of multi-modal datasets, these approaches provide an additional regularization signal that helps prevent the collapse problem and preserves the distinctiveness of learned features.

\subsubsection{Mutual Information Maximization Methods}

Recently, a simple and effective method~\citep{zbontar2021barlow, bardes2021vicreg} based on the information maximization criterion~\citep{linsker1988infomax, thomas2006shannonMI, tschannen2019mutual} has been proposed to address the collapse problem. These methods measure the information between the embedding vectors in the current batch samples, ensuring that the embeddings maintain sufficient variability and do not collapse into a trivial solution. By maximizing mutual information, these approaches promote diversity and preserve the unique features of different input samples, thereby enhancing the overall robustness and effectiveness of the contrastive learning framework.

The design of our DeInfoReg model is inspired by the AL and SCPL models~\citep{wu2021associated, wang21scpl, wu2021associated}. Similarly to AL, the new model addresses the problems of gradient vanishing and exploding in traditional deep neural network models by decomposing the network into multiple blocks with independent optimization objectives. Following SCPL, we enhance the representation learning of an input object by considering its predictive accuracy for the corresponding target and its relationship with other objects. However, DeInfoReg offers significant advantages over these models. Unlike AL, it eliminates the need for multiple fully connected layers near the output layer. Furthermore, unlike SCPL, DeInfoReg does not require paired inputs during training. These simplifications make DeInfoReg more practical and easier to implement in real-world applications. Additionally, we incorporate the concept of information maximization in the design of the loss function~\citep{bardes2021vicreg} to prevent the collapse problem. As a result, the proposed model achieves high accuracy and robustness in various tasks.

\subsection{Model Parallelism, Data Parallelism, and GPipe}

The high computational cost of deep neural network training has led to parallel training research. Among them, data parallelism replicates the entire model on multiple devices and splits the training data into mini-batches~\citep{shallue2019measuring}. Each device computes the gradients independently and the results are later aggregated to update the model parameters. This method is straightforward to implement and works well when the model fits in the memory of a single device.

In contrast, model parallelism divides the neural network across multiple devices, allocating different layers or segments to different processors. This approach is particularly useful when the model is too large to fit in the memory of a single device~\citep{jia2019beyond}. However, traditional model parallelism often faces challenges such as forward and backward locking~\citep{jaderberg2017decoupled}, where the computation in one module must wait for the results from another, thus reducing parallel efficiency.

A notable advancement in model parallelism is GPipe and its variants~\citep{huang2019gpipe,narayanan2019pipedream}, which introduced a pipeline parallelism strategy to alleviate some of these limitations. GPipe subdivides a mini-batch into micro-batches and schedules them sequentially through the pipeline. This overlapping of computations across different segments of the model helps reduce idle times caused by inter-device synchronization, thereby increasing training throughput. Despite these improvements, GPipe continues to experience forward and backward locking challenges, particularly in deep networks, as gradient computations remain constrained by the chain rule.

\section{The Design of DeInfoReg} \label{sec:method}

This section introduces the design of our model, DeInfoReg. We describe the motivation and model architecture, detailing the gradient update mechanism for each module. We discuss the design of local loss functions, their impact, and how to leverage pipelines to achieve model parallelism.

\subsection{Motivation} \label{sec:motivation}

Deep neural networks often suffer from vanishing gradients due to long gradient flows, which hinders effective training. Conventional backpropagation also enforces strict sequential dependencies, limiting opportunities for parallel execution. DeInfoReg decouples a deep network into smaller and independent modules and introduces local loss functions that incorporate variance, invariance, and covariance terms to address these issues. The variance term promotes diversity among embeddings to prevent collapse, the invariance term ensures that embeddings remain consistent with the target labels, and the covariance term reduces redundancy by de-correlating different feature dimensions. This design not only shortens gradient flows, mitigating vanishing gradients, but also enables each module to operate independently, thereby facilitating model parallelism across multiple GPUs and significantly improving training throughput.

\subsection{DeInfoReg Architecture and Gradient Flow Design} \label{chapter:model_and_gradient}

\begin{figure*}[!htb]
    \centering
    \includegraphics[width=1\textwidth]{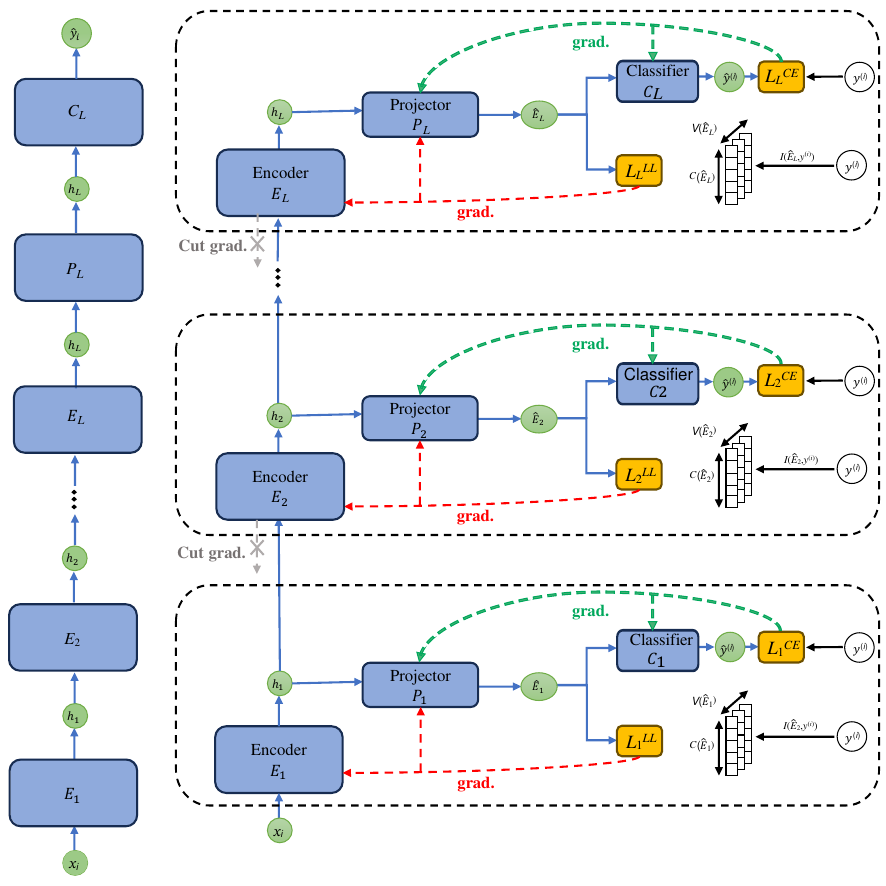}
    \caption{A standard neural network (left) and the DeInfoReg network architecture (right). The green and red dashed arrows indicate areas affected by gradient updates, while gray dashed cross marks indicate gradient truncation.}
    \label{fig:deinforeg-archi}
\end{figure*}

End-to-end backpropagation (BP) is the standard way to compute the gradients of a neural network. However, when a network is deep, the long gradient flow may cause the gradient to vanish or explode, compromising the effectiveness of model training. DeInfoReg decouples a large network into multiple small modules, blocks the gradient flow between modules, and assigns local objectives to each module. Thus, each gradient flow is short and less likely to suffer from the issues of vanishing or exploding gradients. Thus, the training is likely to be more efficient.

We use Figure~\ref{fig:deinforeg-archi} to illustrate the structure design of DeInfoReg and its relationship with a standard neural network. When given a standard neural network to map a feature set $x^{(i)}$ to $\hat{y}^{(i)}$ that approximates the ground truth label $y^{(i)}$ via $L+2$ transformations $E_1, E_2, \ldots, E_L, P_L, C_L$ (each of the $E_i$s, $P_L$, and $C_L$ can be any building block of a graph, such as the fully connected layer, convolution layer, recurrent layer, etc.), we can convert the network into the DeInfoReg architecture with $L$ modules, as shown on the right of Figure~\ref{fig:deinforeg-archi}. We use $P_{l}$, $C_{l}$, and $E_{l}$ to represent the projector, the classifier, and the encoder of the module $l$, respectively; these namings follow the convention in contrastive learning. The input of the encoder $E_{l}$ is the output of the encoder $E_{l-1}$ of the previous module. The output of the encoder ($h_{l}$) is processed by $P_{l}$ to produce the embedding $\hat{E}_{l}$. We update the weights of the encoder $E_l$ and projector $P_l$ to minimize the local loss function $L_l^{LL}$ of layer $l$ (to be introduced in the next section); the red dashed arrows in Figure~\ref{fig:deinforeg-archi} indicate the gradient flow related to the loss $L_l^{LL}$. Furthermore, each module's classifier $C_l$ transforms $\hat{E}_l$ to the predicted label. We update the weights of the projector $P_l$ and the classifier layers $C_l$ to minimize the cross-entropy loss $L_l^{CE}$ between the predicted label $\hat{y}^{(i)}$ and the ground truth label $y^{(i)}$ for layer $l$, as shown by Equation~\ref{eq:loss_ce}.

\begin{equation} \label{eq:loss_ce}
L_l^{CE}(y^{(i)}, \hat{y}^{(i)}) = -\sum_{j=1}^C y^{i,j} \log\left(\hat{y}^{i,j}\right),
\end{equation} 
where $y^{i,j} \in \{0,1\}$ indicates whether instance $i$ belongs to the class $j$, $\hat{y}^{i,j}$ is the predicted probability of class $j$ for instance $i$.

The green dashed arrows in Figure~\ref{fig:deinforeg-archi} show the gradient flow related to the loss $L_l^{CE}$. All the gradients are flowed only within a module.

\subsection{Local Loss Function Design} \label{chapter:local_loss}

\begin{algorithm}[tb]
    \caption{PyTorch pseudocode for the local loss computation for each layer}
    \label{alg:deinforegloss}
    \SetAlgoLined
        \pyth{def forward(self, x, label):}\\
        \Indp
        \pyth{x = self.projector(x)}\\
        \pyth{nor_x =  nn.functional.normalize(x)}\\
        \pyth{batch_size = label.shape[0]}\\
        \pyth{# Variance}\\
        \pyth{x_mean = nor_x - nor_x.mean(dim=0)}\\
        \pyth{std_x = torch.sqrt(x_mean.var(dim=0) + 0.0000001) }\\
        \pyth{var_loss = torch.mean(F.relu(1 - std_x)) / (batch_size)}\\
        \pyth{# Invariance}\\
        \pyth{target_onehot = to_one_hot(label, self.n_class)}\\
        \pyth{target_simi = similarity_matrix(target_onehot)}\\
        \pyth{x_simi = similarity_matrix(nor_x)}\\
        \pyth{invar_loss = F.mse_loss(target_simi, x_simi)}\\
        \pyth{# Covariance}\\
        \pyth{x_mean = nor_x - nor_x.mean(dim=0)}\\
        \pyth{cov_x = (x_mean.T @ x_mean) / (batch_size)}\\
        \pyth{cov_loss = off_diagonal(cov_x).pow(2).sum().div(self.num_features)}\\
        \pyth{ }\\
        \pyth{loss = var_loss + invar_loss + cov_loss}\\
        \pyth{ }\\
        \Indm
    \pyth{# Optimization Step(Update Encoder & Projector)}\\
    \pyth{loss.backward()}\\
    \pyth{optimizer.step()}\\ 
\end{algorithm}

This subsection details the design of the local loss function, which integrates three complementary components: \textit{variance}, \textit{invariance}, and \textit{covariance} losses. Together, these terms ensure the model learns meaningful representations while avoiding collapse -- a common issue in self-supervised learning~\citep{bardes2021vicreg}. 

\paragraph{Variance Loss: Promoting Diversity}

The \textit{variance loss} is designed to promote diversity among embeddings within a batch, which is crucial for preventing the collapse problem. Collapse occurs when embeddings in a batch become overly similar, resulting in a loss of representational capacity. The variance loss addresses this by penalizing dimensions with low variability across embeddings, ensuring sufficient diversity.

Given a batch of embeddings $\mathcal{E}_l \in R^{N \times C}$ for a layer $l$, where $N$ is the number of samples and $C$ is the dimensionality of each embedding, we first compute the vector $\bar{e}$, which is the average of the rows in $\mathcal{E}_l$:

\begin{equation} \label{eq:column_averaged_embedding} 
\bar{e}_l = [\bar{e}_l^1, \ldots, \bar{e}_l^C] = \left[\frac{1}{N}\sum_{i=1}^N e_l^{i,1}, \ldots, \frac{1}{N}\sum_{i=1}^N e_l^{i,N}\right],
\end{equation}
where $e_l^{i,j}$ denotes the $(i,j)$th entry of $\mathcal{E}_l$.

Next, we compute the centered embedding matrix $\hat{E}_l$ by subtracting the column mean $\bar{e}_l$: 

\begin{equation} \label{eq:centered_embedding} 
\hat{E}_l = \mathcal{E}_l - \bar{e}_l = \left[e_l^{i,j} - \bar{e}_l^i\right]_{i=1,\ldots,N, j=1,\ldots,C}. 
\end{equation}

The variance loss is then defined as: 

\begin{equation} \label{eq:variance_loss} 
V(\hat{E}_l) = \frac{1}{C} \max\left(0, \gamma - S(\hat{E}_l)\right), 
\end{equation} 
where $S(\hat{E}_l) = [s_l^1, \ldots, s_l^C]$, each $s_l^i$ is the standard deviation of the $i$th column in $\hat{E}_l$, and $\gamma$ is a predefined threshold. The hinge function $\max(0, \gamma-S(\hat{E}_l))$
penalizes dimensions with insufficient variability.

\paragraph{Invariance Loss: Preserving Consistency}

The \textit{invariance loss} ensures that the learned embeddings $e_l^i$ are consistent with the target labels $y^{(i)}$. Let $Y = [y^{i,j}] \in \{0, 1\}^{N \times C}$ represent the ground-truth labels for a batch. The invariance loss $I(\hat{E}_l, Y)$ is computed as:

\begin{equation} \label{eq:inv-loss}
I(\hat{E}_l, Y) = \frac{1}{N} \left\|Sim(Y) - Sim(Norm(\hat{E}_l))\right\|^2,
\end{equation}
where $Norm(X) = \bar{X} / (\left\|\bar{X}\right\|_2 + \epsilon)$, $\bar{X}$ is a row-centered matrix: $\bar{X} = [\bar{x}_{i,j}] = [x_{i,j} - \sum_{j=1}^C x_{i,j} / C]$, $\left\|\bar{X}\right\|_2$ returns the L2 norm of each instance (i.e., each row), $Sim(X) = XX^T $ where each entry at the $i$th row and $j$th column in the matrix represents the cosine similarity between row $i$ and row $j$ in $X$, and $\epsilon$ is a small constant added to prevent division by zero.  This loss aligns the embedding similarity structure with that of the normalized label vectors, enhancing semantic consistency.

\paragraph{Covariance Loss: Reducing Redundancy}

The \textit{covariance loss} minimizes redundancy by de-correlating different dimensions of the embeddings. The covariance matrix $Cov(\hat{E}_l)$ is given by:

\begin{equation} \label{eq:cov-mat}
   Cov(\hat{E}_l) = \frac{1}{N} (\hat{E}_l - \tilde{e}_l)^T(\hat{E}_l - \tilde{e}_l),
\end{equation}
where $\tilde{e}_l=\sum_{i=1}^N \hat{e}_l^i / N$ is the average of row vectors in the matrix $\hat{E}_l$ ($\hat{e}_l^i$ is the $i$th row in $\hat{E}_l$).

The covariance loss $Cov(\hat{E}_l)$ penalizes off-diagonal entries of the covariance matrix:

\begin{equation}  \label{eq:covar-loss}
C(\hat{E}_l) = \frac{1}{C}\sum_{i\neq j}[Cov(\hat{E}_l)]_{i,j}^{2},
\end{equation}

This de-correlation reduces redundancies in the embeddings, encouraging efficient representation learning.

The final local loss function is the sum of variance, invariance, and covariance losses, as defined in Equation~\ref{eq:local_loss}, and a pseudocode to compute the local loss is given in Algorithm~\ref{alg:deinforegloss}.

\begin{equation} 
\label{eq:local_loss}
\begin{aligned}
L^{LL}_l(\hat{E}_l, Y) = V(\hat{E}_l) + I(\hat{E}_l, Y) + C(\hat{E}_l).
\end{aligned}
\end{equation}

The final loss function in each module, given by Equation~\ref{eq:loss_total}, includes both the cross-entropy loss $L_l^{CE}$ (Equation~\ref{eq:loss_ce}) and the local loss $L_l^{LL}$ (Equation~\ref{eq:local_loss}) for layer $l$. The hyperparameter $\alpha$ is set to 0.001. In other words, local loss plays a much more important role in training.

\begin{equation} 
\label{eq:loss_total}
L =  \sum_{l=1}^{L}\left ( L_l^{LL} + \alpha  L_l^{CE} \right ),
\end{equation} 
where $L$ is the number of layers.

\subsection{Integrating with the Pipeline Strategy}

The decoupled nature of the DeInfoReg framework not only mitigates the issues of gradient vanishing but also facilitates the way for efficient model parallelism through a pipeline strategy. This section details how DeInfoReg can be integrated with a pipeline approach to further enhance training throughput and resource utilization.

\subsubsection{Overview}

Traditional end-to-end backpropagation suffers from both forward and backward locking, where computations across layers are strictly sequential. In contrast, DeInfoReg decomposes a deep neural network into multiple independent modules, each equipped with its own local loss function (comprising variance, invariance, and covariance components) and a dedicated classifier. Since the gradient flow is truncated at the module boundaries, the inter-module dependencies are minimized. This modularity is a natural fit for a pipeline strategy, where different modules can be distributed across multiple processing units (e.g., GPUs) and processed concurrently.

\subsubsection{Pipeline Implementation in DeInfoReg}

In DeInfoReg, each component is designed with its own local loss function and operates independently. As soon as the forward pass of the preceding layers is completed, the next component is triggered to start its forward-backward process. This means that once the necessary outputs from previous layers are available, a component starts processing its input through a forward pass and then computes its local loss, triggering its backward pass.

\begin{figure*}[!htb]
    \centering
    \includegraphics[width=1\textwidth]{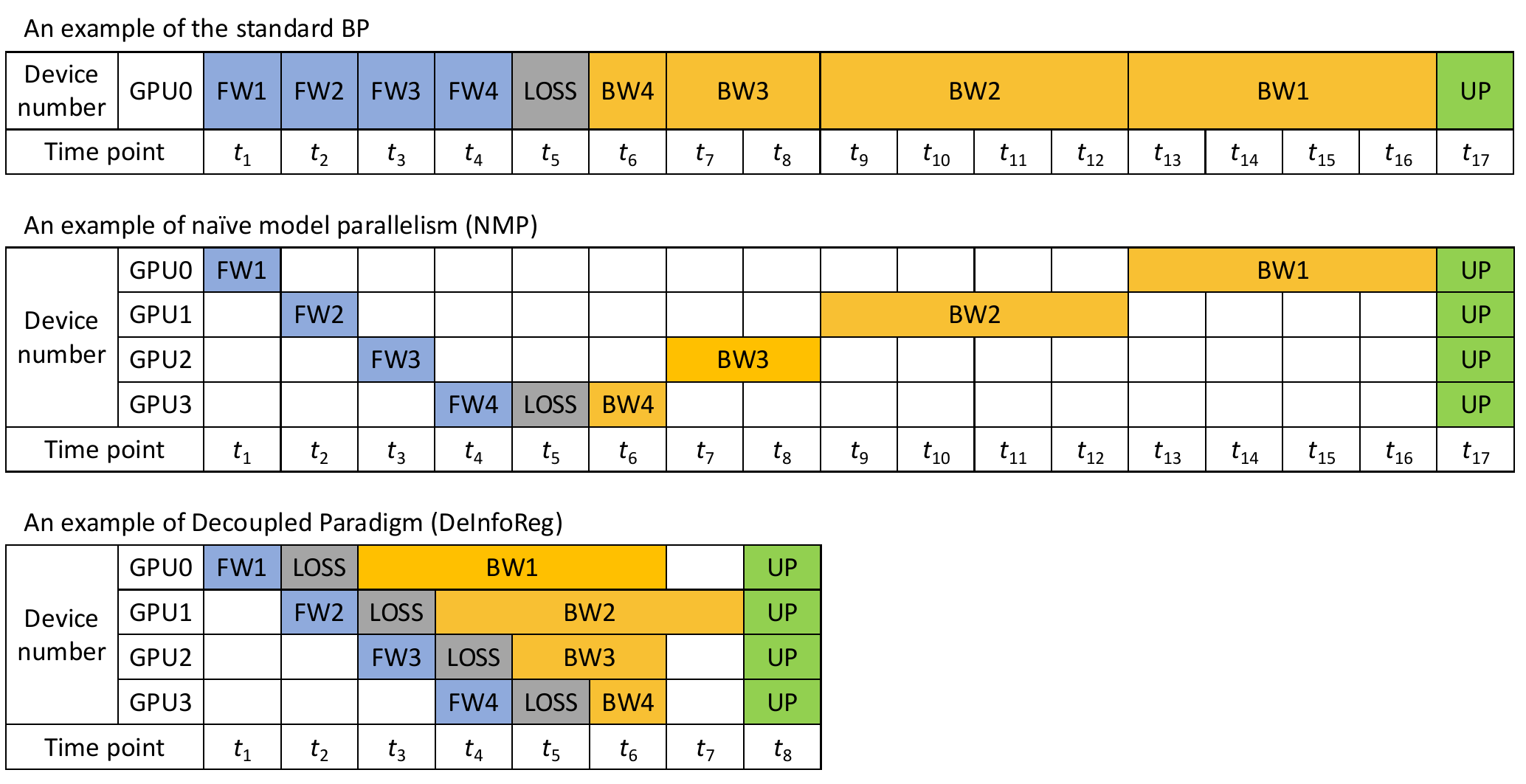}
    \caption{An illustrative example comparing the GPU usage per iteration for standard BP, na\"{i}ve model parallelism (NMP), and DeInfoReg. FW$i$ and BW$i$ denote the forward and backward of component $i$, LOSS is loss computation, and UP is parameter update.}
    \label{fig:pipeline-deinforeg}
\end{figure*}

Figure~\ref{fig:pipeline-deinforeg} illustrates the process of applying pipelining to DeInfoReg. Suppose that we divide a neural network into four components. With standard backpropagation, as shown at the top of the figure, each forward operation (FW$i$) must wait for the completion of all preceding forward operations (i.e., FW$1$, $\ldots$, FW$i-1$), and each backward operation (BW$i$) must wait for the backward operations of subsequent layers (i.e., BW$i+1$, $\ldots$, BW$L$).

If we na\"{i}vely allocate the four components to four processing units, such as GPUs (as shown in the middle of Figure~\ref{fig:pipeline-deinforeg}), the inherent dependencies between forward and backward operations remain unchanged, causing all operations to execute sequentially. 

When using DeInfoReg, as depicted at the bottom of the figure, once the first GPU (GPU0) completes its forward operation, the second GPU (GPU1) can immediately begin processing the next component using the output from GPU0. Although forward operations remain dependent, backward operations can be executed concurrently. This setup enables multiple GPUs to operate in parallel, significantly boosting computational efficiency. In this example, the training time per iteration is reduced from 17 to 8 time units.

\subsection{Comparison with Existing Approaches}

Although variance, invariance, and covariance losses have been explored in prior work~\citep{bardes2021vicreg}, our contribution lies in integrating supervised contrastive learning within a decoupled paradigm. Existing methods typically apply these losses within conventional end-to-end frameworks, which still suffer from long gradient flows and strict sequential dependencies. In contrast, our approach leverages the strengths of supervised contrastive learning to guide each module's representation learning while decoupling the network into independent units. This integration not only shortens gradient paths, reducing the risk of vanishing gradients, but also enables efficient model parallelism across multiple GPUs.

\section{Experiments} \label{sec:exp}

\begin{table}[tb]
    \centering
    \caption{Accuracy for BP, AL, SCPL, and DeInfoReg across NLP tasks with LSTM under different batch sizes}
    \label{tab:LSTM_acc}
    \begin{tabular}{c|cccc}
    \toprule
    Batch Size       & 64         & 128         & 256         & 512         \\ 
    \midrule
    \multicolumn{5}{c}{\textbf{IMDB}} \\ 
    \midrule
    BP                   & 88.52 $\pm$ 0.43    & 88.17 $\pm$ 0.41     & 87.82 $\pm$ 0.68     & 86.98 $\pm$ 0.47     \\
    AL                   & 89.53 $\pm$ 0.37    & 89.77 $\pm$ 0.26     & 89.8 $\pm$ 0.1       & 89.58 $\pm$ 0.23     \\
    SCPL                 & 89.59 $\pm$ 0.46    & 89.48 $\pm$ 0.24     & 89.64 $\pm$ 0.22     & 89.81 $\pm$ 0.32     \\
    DeInfoReg            & \textbf{89.83 $\pm$ 0.48} & \textbf{90.13 $\pm$ 0.06} & \textbf{90.1 $\pm$ 0.22} & \textbf{90.02 $\pm$ 0.31} \\
    \midrule
    \multicolumn{5}{c}{\textbf{AGNews}} \\ 
    \midrule
    BP                   & 91.56 $\pm$ 0.16    & 91.49 $\pm$ 0.08     & 91.47 $\pm$ 0.11     & 91.25 $\pm$ 0.07     \\
    AL                   & 92.16 $\pm$ 0.08    & 92.12 $\pm$ 0.03     & 92.07 $\pm$ 0.11     & 91.98 $\pm$ 0.12     \\
    SCPL                 & 91.92 $\pm$ 0.14    & 91.77 $\pm$ 0.1      & 92.11 $\pm$ 0.16     & 92.09 $\pm$ 0.11     \\
    DeInfoReg            & \textbf{92.26 $\pm$ 0.12} & \textbf{92.3 $\pm$ 0.06}  & \textbf{92.22 $\pm$ 0.1}  & \textbf{92.21 $\pm$ 0.09}  \\
    \midrule
    \multicolumn{5}{c}{\textbf{DBpedia}} \\ 
    \midrule
    BP                   & 98.56 $\pm$ 0.05    & 98.51 $\pm$ 0.04     & 98.48 $\pm$ 0.03     & 98.42 $\pm$ 0.02     \\
    AL                   & 98.57 $\pm$ 0.06    & 98.57 $\pm$ 0.06     & 98.59 $\pm$ 0.03     & 98.56 $\pm$ 0.01     \\
    SCPL                 & 98.56 $\pm$ 0.04    & 98.57 $\pm$ 0.06     & 98.56 $\pm$ 0.01     & 98.58 $\pm$ 0.02     \\
    DeInfoReg            & \textbf{98.58 $\pm$ 0.03} & \textbf{98.57 $\pm$ 0.01} & \textbf{98.62 $\pm$ 0.01} & \textbf{98.65 $\pm$ 0.01} \\
    \bottomrule
    \end{tabular}
\end{table}

\begin{table}[tbh]
    \centering
    \caption{Accuracy for BP, AL, SCPL, and DeInfoReg across NLP datasets with Transformer under different batch sizes}
    \label{tab:Transformer_acc}
    \begin{tabular}{c|cccc}
    \toprule
    Batch Size & 64                    & 128                   & 256                   \\ 
    \midrule
    \multicolumn{4}{c}{\textbf{IMDB}} \\ 
    \midrule
    BP                & 87.91 $\pm$ 0.44      & 87.54 $\pm$ 0.29      & 87.50 $\pm$ 0.42      \\
    AL                & 87.93 $\pm$ 0.37      & 86.89 $\pm$ 0.73      & 87.98 $\pm$ 0.48      \\
    SCPL              & 88.16 $\pm$ 0.48      & 88.89 $\pm$ 0.29      & 88.69 $\pm$ 0.32      \\
    DeInfoReg         & \textbf{89.02 $\pm$ 0.17} & \textbf{89.26 $\pm$ 0.09} & \textbf{88.86 $\pm$ 0.30} \\
    \midrule
    \multicolumn{4}{c}{\textbf{AGNews}} \\ 
    \midrule
    BP                & 91.05 $\pm$ 0.10      & 90.84 $\pm$ 0.11      & 90.74 $\pm$ 0.16      \\
    AL                & 88.20 $\pm$ 1.60      & 86.20 $\pm$ 1.90      & 89.41 $\pm$ 0.86      \\
    SCPL              & 91.36 $\pm$ 0.21      & 91.64 $\pm$ 0.07      & 91.66 $\pm$ 0.21      \\
    DeInfoReg         & \textbf{91.91 $\pm$ 0.13} & \textbf{91.95 $\pm$ 0.06} & \textbf{91.90 $\pm$ 0.15} \\
    \midrule
    \multicolumn{4}{c}{\textbf{DBpedia}} \\ 
    \midrule
    BP                & 97.66 $\pm$ 0.05      & 97.63 $\pm$ 0.01      & 97.59 $\pm$ 0.02      \\
    AL                & 91.20 $\pm$ 2.37      & 92.81 $\pm$ 2.02      & 93.58 $\pm$ 1.62      \\
    SCPL              & 97.35 $\pm$ 0.27      & 97.47 $\pm$ 0.04      & 97.58 $\pm$ 0.05      \\
    DeInfoReg         & \textbf{97.84 $\pm$ 0.02} & \textbf{97.84 $\pm$ 0.02} & \textbf{97.85 $\pm$ 0.03} \\
    \bottomrule
    \end{tabular}
\end{table}

This section presents the experimental setups and results. We conduct experiments to compare DeInfoReg with backpropagation (BP) and other decoupled models Associated Learning (AL) and Supervised Contrastive Parallel Learning (SCPL) using several network architectures, including Long Short-Term Memory (LSTM), Transformer, VGG, and ResNet, based on open datasets in natural language processing and computer vision. Experimental results show that DeInfoReg outperforms the baselines on most datasets and most network architectures (Section~\ref{sec:nlp-exp} and Section~\ref{sec:cv-exp}).

We conduct experiments to study possible reasons for DeInfoReg's surprisingly great prediction accuracy. First, we investigate whether short gradient flows alleviate the gradient vanishing issue (Section~\ref{sec:gradient-exp}). Specifically, we demonstrate that: (1) deep neural networks are easier to train with DeInfoReg compared to BP, as evidenced by higher prediction accuracies, and (2) the gradient magnitudes near the input layer of a deep neural network approach zero when trained with BP, whereas they remain significantly larger when trained with DeInfoReg. This observation suggests that shorter gradient flows facilitate more effective signal transmission. Additionally, we show that DeInfoReg is more robust when it comes to noisy labels than BP (Section~\ref{sec:noisy-label-exp}). 

In addition to the accuracy evaluations, this work also investigates practical aspects of deploying DeInfoReg. We present an analysis of the training time speedup achieved by integrating DeInfoReg with a pipeline strategy for model parallelism (Section~\ref{sec:mp-exp}). Furthermore, an empirical examination of the model's memory footprint is provided (Section~\ref{sec:memory-exp}).

Finally, we conduct an ablation study on the local loss functions and a hyperparameter analysis on the value of $\alpha$ in Equation~\ref{eq:loss_total}. It turns out that all three local losses are essential (Section~\ref{sec:loss-func-exp}), and the value of $\alpha$ appears to be optimal around 1e-2 to 1e-3 (Section~\ref{sec:alpha-value-exp}). 

All experiments are conducted on a machine with an Intel Core i7-12700K CPU (3.60 GHz, 25MB cache) and an NVIDIA GeForce RTX 3090 GPU (24GB). We perform a hyperparameter search for each model using the validation dataset to ensure that each model is optimally tuned for fair and accurate comparison. The hyperparameter settings can be found in the released code.

\subsection{Natural Language Processing Experiments} \label{sec:nlp-exp}

\subsubsection{Experiment Setup} 
We use two network structures that are commonly used for NLP experiments, LSTM and Transformer. We use three open datasets: IMDB, AGNews, and DBpedia.

In NLP experiments, we remove blank words and stop words. The first layer of each model is a pre-trained GloVe (Global Vectors for Word Representation)~\citep{pennington2014glove}. DeInfoReg and SCPL use the identity function as the projector in each layer. The hyperparameters of SCPL follow the original paper~\citep{wang21scpl} with a temperature parameter of 0.1.

The LSTM network comprises the main network encoder plus two fully connected layers as the classifier. We separate these components into four modules. DeInfoReg, SCPL, and AL block the gradients between the four modules. Each module includes four layers of bidirectional LSTM with a dimension size of 300. SCPL and BP add the classifier after the output of the last module with two fully connected layers and tanh as the activation function for the first linear layer. For DeInfoReg, the structure of the classifier in each module includes a tanh activation function followed by a fully connected layer.

The Transformer network comprises four modules comprising the main network encoder and two fully connected layers forming the classifier. DeInfoReg, SCPL, and AL block the gradient flows among the four modules. Each module includes three Transformer encoding layers, each of which includes a Multi-Head Self-Attention layer, an add-and-norm layer (i.e., a residual connection and layer normalization), a fully connected layer, and another add-and-norm layer. SCPL and BP add a classifier after the output of the last module with two linear layers where the activation function of the first layer is tanh. The classifier for each module in DeInfoReg is a tanh function followed by a fully connected layer.

\subsubsection{NLP Classification Results}

We test the LSTM network with batch sizes of 64, 128, 256, and 512. We test different batch sizes because models based on contrastive learning often require a larger batch size to get decent results. Similarly, we test batch sizes of 64, 128, and 256 for the Transformer network. We do not test the batch size 512 for Transformer due to GPU memory limitations. 

The results for LSTM and Transformer deep networks are shown in Table~\ref{tab:LSTM_acc} and Table~\ref{tab:Transformer_acc}, respectively. We repeat each experiment five times and report the mean $\pm$ standard deviation of accuracy. The results show that DeInfoReg achieves the best accuracy in all datasets in all batch sizes compared to BP, AL, and SCPL. In addition, batch size has a limited influence on the accuracy of DeInfoReg.

\subsection{Computer Vision Experiments} \label{sec:cv-exp}
\subsubsection{Experiment Setup}

We use two network structures that are widely used for computer vision tasks: VGG network (VGG-11)~\citep{simonyan2014very} and ResNet (ResNet-18)~\citep{he2016deep}. We use three open datasets: CIFAR-10, CIFAR-100, and TinyImageNet. We scale the TinyImageNet images from $64 \times 64 \times 3$ to $32 \times 32 \times 3$ to match the input dimension of the network.

We augment input images (e.g., random cropping, random flipping, color jittering, and random grayscale) as a preprocessing step for all models. SCPL and DeInfoReg use a multilayer perceptron as the projector layer. In particular, SCPL follows the settings in the original paper~\citep{wang21scpl}: two fully connected layers with ReLU as the activation function for the first layer. As for DeInfoReg, the projector includes three fully connected layers, where the output of the first two layers also goes through batch normalization and the ReLU activation function.

The VGG network comprises nine convolutional layers and four max-pooling layers as the main network encoder, and two fully connected layers as the classifier. Each convolutional layer uses batch normalization and ReLU as the activation function~\citep{ioffe2015batch}. In DeInfoReg, SCPL, and AL, the main network encoder is divided into four modules; the gradient flow between modules is blocked. Each module consists of four convolution layers, followed by a max-pooling layer, then two convolution layers and another max-pooling layer, followed by two more convolution layers and a third max-pooling layer, and finally, one convolution layer and a max-pooling layer. SCPL and BP add a classifier after the last module; the structure of the classifier is two fully connected layers, where ReLU is the activation function of the first fully connected layer. DeInfoReg includes a classifier for each block. The structure of each classifier includes batch normalization, a ReLU activation function, and a fully connected layer.

The ResNet comprises 18 residual layers as the main network encoder, with one average pooling layer and one fully connected layer as the classifier. In DeInfoReg, SCPL, and AL, the main network encoder is divided into four modules with gradient truncation. Each module includes eight blocks; each block consists of two convolutional layers with a residual connection. BP and SCPL add a classifier after the last module, structured as an average pooling layer followed by a fully connected layer. DeInfoReg includes a classifier for each block; the classifier is structured as a batch normalization layer, followed by a ReLU activation function and a fully connected layer.

\subsubsection{Computer Vision Classification Results} \label{vision_Experiment}

\begin{table}[tbh]
    \centering
    \caption{Accuracy for BP, AL, SCPL, and DeInfoReg across CV tasks with VGG under different batch sizes}
    \label{tab:VGG_acc}
    \begin{tabular}{c|cccc}
    \toprule
    Batch Size & 64                    & 128                   & 256                   & 512                   \\ 
    \midrule
    \multicolumn{5}{c}{\textbf{CIFAR-10}} \\ 
    \midrule
    BP                & \textbf{93.45 $\pm$ 0.09} & \textbf{93.71 $\pm$ 0.16} & \textbf{93.63 $\pm$ 0.1}  & \textbf{93.74 $\pm$ 0.24}  \\
    AL                & 93.02 $\pm$ 0.18          & 92.77 $\pm$ 0.17          & 92.45 $\pm$ 0.21          & 92.23 $\pm$ 0.34           \\
    SCPL              & 93.34 $\pm$ 0.05          & 93.41 $\pm$ 0.19          & 93.6 $\pm$ 0.14           & 93.67 $\pm$ 0.24           \\
    DeInfoReg         & 93.02 $\pm$ 0.18          & 93.00 $\pm$ 0.14          & 93.22 $\pm$ 0.1           & 93.1 $\pm$ 0.05            \\ 
    \midrule
    \multicolumn{5}{c}{\textbf{CIFAR-100}} \\ 
    \midrule
    BP                & 68.13 $\pm$ 0.84          & 69.77 $\pm$ 0.54          & 71.02 $\pm$ 0.17          & 71.77 $\pm$ 0.34           \\
    AL                & 70.47 $\pm$ 0.24          & 69.91 $\pm$ 0.23          & 69.73 $\pm$ 0.19          & 69.1 $\pm$ 0.22            \\
    SCPL              & 70.19 $\pm$ 0.32          & 71.42 $\pm$ 0.19          & 71.85 $\pm$ 0.17          & 72.33 $\pm$ 0.19           \\
    DeInfoReg         & \textbf{73.7 $\pm$ 0.17}  & \textbf{73.67 $\pm$ 0.24} & \textbf{74.03 $\pm$ 0.15} & \textbf{74.18 $\pm$ 0.12}  \\ 
    \midrule
    \multicolumn{5}{c}{\textbf{TinyImageNet}} \\ 
    \midrule
    BP                & 45.83 $\pm$ 0.66          & 42.77 $\pm$ 0.13          & 43.7  $\pm$ 1.1           & 45.52 $\pm$ 0.73           \\
    AL                & 46.73 $\pm$ 0.16          & 46.53 $\pm$ 0.27          & 46.8 $\pm$ 0.23           & 46.19 $\pm$ 0.19           \\
    SCPL              & 43.7 $\pm$ 0.77           & 44.71 $\pm$ 0.3           & 45.1 $\pm$ 0.3            & 45.58 $\pm$ 0.35           \\
    DeInfoReg         & \textbf{49.74 $\pm$ 0.32} & \textbf{50.93 $\pm$ 0.36} & \textbf{51.48 $\pm$ 0.56} & \textbf{50.98 $\pm$ 0.26}  \\
    \bottomrule
    \end{tabular}
\end{table}

\begin{table}[tbh]
    \centering
    \caption{Accuracy for BP, AL, SCPL, and DeInfoReg across CV tasks with ResNet18 under different batch sizes}
    \label{tab:resnet18_acc}
    \begin{tabular}{c|cccc}
    \toprule
    Batch Size & 64                    & 128                   & 256                   & 512                   \\ 
    \midrule
    \multicolumn{5}{c}{\textbf{CIFAR-10}} \\ 
    \midrule
    BP                & \textbf{93.4 $\pm$ 0.27}  & \textbf{93.47 $\pm$ 0.14} & \textbf{93.48 $\pm$ 0.13} & \textbf{93.55 $\pm$ 0.12}  \\
    AL                & 90.63 $\pm$ 0.28          & 90.63 $\pm$ 0.13          & 90.35 $\pm$ 0.14          & 89.82 $\pm$ 0.18           \\
    SCPL              & 92.09 $\pm$ 0.10          & 92.33 $\pm$ 0.11          & 92.24 $\pm$ 0.12          & 92.00 $\pm$ 0.19           \\
    DeInfoReg         & 91.77 $\pm$ 0.10          & 92.03 $\pm$ 0.19          & 92.01 $\pm$ 0.12          & 91.96 $\pm$ 0.19           \\ 
    \midrule
    \multicolumn{5}{c}{\textbf{CIFAR-100}} \\ 
    \midrule
    BP                & 70.58 $\pm$ 0.06          & 70.67 $\pm$ 0.12          & 71.00 $\pm$ 0.38          & 70.97 $\pm$ 0.34           \\
    AL                & 70.47 $\pm$ 0.24          & 69.91 $\pm$ 0.23          & 69.73 $\pm$ 0.19          & 69.10 $\pm$ 0.22           \\
    SCPL              & 67.68 $\pm$ 0.15          & 68.56 $\pm$ 0.41          & 68.95 $\pm$ 0.11          & 69.00 $\pm$ 0.18           \\
    DeInfoReg         & \textbf{70.65 $\pm$ 0.11} & \textbf{70.75 $\pm$ 0.23} & \textbf{71.44 $\pm$ 0.25} & \textbf{71.56 $\pm$ 0.17}  \\ 
    \midrule
    \multicolumn{5}{c}{\textbf{TinyImageNet}} \\ 
    \midrule
    BP                & 47.66 $\pm$ 0.26          & 47.62 $\pm$ 0.20          & 47.73 $\pm$ 0.13          & 47.63 $\pm$ 0.15           \\
    AL                & 42.53 $\pm$ 1.33          & 42.58 $\pm$ 0.44          & 42.52 $\pm$ 0.47          & 41.93 $\pm$ 0.25           \\
    SCPL              & 41.33 $\pm$ 0.42          & 43.69 $\pm$ 0.11          & 44.85 $\pm$ 0.30          & 46.06 $\pm$ 0.40           \\
    DeInfoReg         & \textbf{47.68 $\pm$ 0.14} & \textbf{47.85 $\pm$ 0.35} & \textbf{48.49 $\pm$ 0.32} & \textbf{49.22 $\pm$ 0.20}  \\
    \bottomrule
    \end{tabular}
\end{table}

We train all models with 200 epochs in computer vision classification tasks. Similar to the experiments in NLP, we test different batch sizes: 64, 128, 256, and 512.

The results of the VGG and ResNet models are shown in Table~\ref{tab:VGG_acc} and Table~\ref{tab:resnet18_acc}, respectively. BP performs best on the simplest CIFAR-10 dataset, and DeInfoReg performs best on more complicated CIFAR-100 and TinyImageNet on both VGG and ResNet networks. Additionally, DeInfoReg shows stable accuracy across different batch sizes. This discovery indicates that we can use a small batch size to save GPU memory for DeInfoReg. For comparison, SCPL is also based on contrastive learning, but SCPL usually requires a larger batch size to get better results. DeInfoReg has a smaller request on batch size, likely because its loss function considers the variance and covariance regularization directly, so negative pairs are less critical.

\subsection{DeInfoReg Mitigates Gradient Vanishing} \label{sec:gradient-exp}

\begin{table}[tbh]
    \centering
    \caption{Test accuracy of BP and DeInfoReg at various module depths $L$}
    \label{tab:gradient-magnitude}
    \resizebox{\columnwidth}{!}{
    \begin{tabular}{c|cccccc} 
    \toprule
    $L$ & 4     & 6     & 8     & 10    & 12    & 14     \\ 
    \hline\hline
    BP            & $91.08 \pm 0.21$ & $91.55 \pm 0.07$ & $91.41 \pm 0.30$ & $50.78 \pm 0.16$ & $45.14 \pm 0.23$ & $25 \pm 0.24$     \\
    DeInfoReg     & $92.03 \pm 0.18$ & $92.2 \pm 0.07$ & $92.11 \pm 0.06$ & $92.1 \pm 0.16$ & $92.28 \pm 0.28$ & $92.12 \pm 0.29$  \\
    \bottomrule
    \end{tabular}
    }
\end{table}

\begin{figure}[tbh]
    \centering
    \includegraphics[width=1\columnwidth]{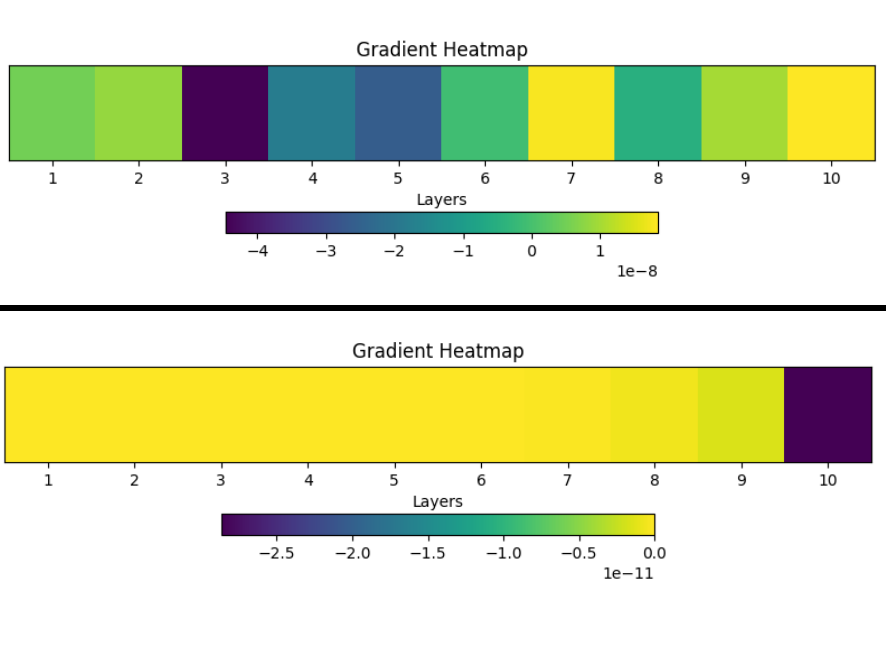}
    \caption{Gradient changes in the encoder output of 10 LSTM blocks (Top: DeInfoReg, Bottom: BP)}
    \label{figure:10_layer_gradient}
\end{figure}

This section experiments on whether short gradient flows mitigate the gradient vanishing issue.

We perform experiments using the LSTM network based on the AGNews dataset. We test the number of modules as 4, 6, 8, 10, 12, and 14. We compare DeInfoReg and BP's accuracies and the magnitude of the gradient values in different layers.

Table~\ref{tab:gradient-magnitude} gives the comparison of the test accuracy of BP and DeInfoReg with different numbers of modules. The results show that BP and DeInfoReg perform well, up to 8 LSTM blocks. However, when the model has ten or more blocks, BP's accuracy drops significantly, while DeInfoReg maintains stable performance, demonstrating better stability and being less sensitive to the number of layers.

Figure~\ref{figure:10_layer_gradient} shows the average magnitudes of the gradient in the encoders of a 10-module (i.e., 10-layer) LSTM network, where the first module is closest to the input, and the tenth module is closest to the output. When trained with BP, the magnitudes of the gradient in the modules near the input approach zero, leading to inefficient training. In contrast, using DeInfoReg maintains reasonable gradient magnitudes near the input. This probably explains why training a deep network with BP often results in poor accuracy, as shown in Table~\ref{tab:gradient-magnitude}. Note that the gradient measurements for DeInfoReg only consider the encoders $E_1, \ldots, E_L$ depicted in Figure~\ref{fig:deinforeg-archi}, excluding gradients from other components such as projectors $P_1, \ldots, P_L$. This approach guarantees that the comparison with the gradients in BP layers is fair and isolates the effect of our method on the gradient flow. The results clearly demonstrate that the increase in gradient magnitude is attributable to alleviating the vanishing gradient problem rather than merely the result of additional loss supervision or increased loss weight.

Overall, these results indicate that DeInfoReg performs more stably than BP at different depths, especially in deeper structures, proving its effectiveness in handling the gradient vanishing issue.

\subsection{Robustness on Noisy Labels} \label{sec:noisy-label-exp}

\begin{table*}[tbh]
    \centering
    \caption{Test accuracy of BP and DeInfoReg under different noisy label ratios $\theta$}
    \label{tab:noisy-ratio-acc}
        \begin{tabular}{c|ccccc} 
        \toprule
        $\theta$ & 0.0          & 0.2          & 0.4          & 0.6          & 0.8             \\ 
        \hline\hline
        BP                                 & 88.27 $\pm$ 0.48 & 83.01 $\pm$ 1.2  & 76.08 $\pm$ 0.13 & 67.71 $\pm$ 0.4  & 57.94 $\pm$ 1.75  \\
        DeInfoReg                          & 90.15 $\pm$ 0.18 & 85.46 $\pm$ 0.66 & 81.28 $\pm$ 1.1  & 76.36 $\pm$ 5.18 & 68.1 $\pm$ 8.31   \\
        \bottomrule
        \end{tabular}
\end{table*}

This section reports DeInfoReg's ability to handle noisy (i.e., incorrect) labels during training.

We compare the accuracy of BP and DeInfoReg's robustness on noisy datasets using the IMDB dataset. We introduce random noisy labels with different noisy ratios into the training data. We test the models' accuracies with clean labels based on the test dataset. 

Table~\ref{tab:noisy-ratio-acc} presents the test accuracy in various noisy ratios. The results indicate that DeInfoReg consistently outperforms BP at all levels of noise. When the noise ratio is high, the performance gap between DeInfoReg and BP widens, suggesting that DeInfoReg's robustness to noise is significantly superior. This increased resilience to noisy labels highlights the effectiveness of our method in preserving meaningful patterns in the data, even under challenging conditions.

\subsection{Training Time Speedup from Model Parallelism} \label{sec:mp-exp}

\begin{table}[tb]
\centering
\caption{The speedup of the training time (using BP as the reference).}
\label{tab:speedup-cmp}
\begin{tabular}{@{}c||ccc@{}}
\toprule
Batch size         & 256            & 512            & 1024          \\ \midrule
BP                 & 1x (28.74 sec) & 1x (28.44 sec) & 1x (28.8 sec) \\
DeInfoReg (1 GPU)  & 0.81x          & 0.85x          & 0.90x         \\
DeInfoReg (2 GPUs) & 1.30x          & 1.35x          & 1.35x         \\
DeInfoReg (4 GPUs) & 1.41x          & 1.44x          & 1.47x         \\ \bottomrule
\end{tabular}
\end{table}

The experimental results presented in Table~\ref{tab:speedup-cmp} illustrate the impact of model parallelism on the training time of DeInfoReg relative to standard backpropagation. We make comparisons in various configurations using one, two, and four GPUs. The experiments were conducted on a VGG model trained on the CIFAR-100 dataset for 100 epochs, and the average training time per epoch was recorded. When using BP as the baseline, training time remains relatively constant across different batch sizes, averaging around 28 -- 29 seconds. In contrast, DeInfoReg exhibits varying speedups depending on the number of GPUs employed.

For instance, with a single GPU, DeInfoReg achieves a speedup factor of less than 1.0 across all batch sizes, indicating that the communication cost between GPUs and the overhead introduced by the decoupled processing in DeInfoReg slightly increases training time compared to BP in this scenario. However, as additional GPUs are leveraged, the benefits of model parallelism become evident. When using two GPUs, the training time is reduced by approximately $30\% -- 35\%$ relative to BP. DeInfoReg further improves the speedup with four GPUs, reaching improvements of $41\%$ to $47\%$.

These results underscore the effectiveness of DeInfoReg's decoupled learning strategy when combined with model parallelism. By partitioning the network into independent components that can perform forward and backward passes concurrently across multiple GPUs, DeInfoReg alleviates the constraints of traditional end-to-end backpropagation and significantly enhances training throughput. This efficiency gain becomes particularly pronounced as more GPUs are employed, demonstrating that DeInfoReg is well suited for scaling up training in environments with abundant parallel computing resources.

\subsection{Comparison of GPU Memory Usage} \label{sec:memory-exp}

\begin{table}[tbh]
\centering
\caption{Comparison of training parameter counts and GPU memory usage for BP, AL, SCPL, and DeInfoReg on an IMDB natural language task (batch size 256); the backbone network is LSTM.}
\label{table:NLP_params}
    \begin{tabular}{cc} 
    \toprule
    Method & GPU Memory Usage  \\ 
    \hline
    BP & 18088 Mib \\
    AL & 18116 MiB \\
    SCPL & 18184 MiB \\
    DeInfoReg & 18104 MiB \\
    \bottomrule
    \end{tabular}
\end{table}

Table~\ref{table:NLP_params} compares the GPU memory usage for BP, AL, SCPL, and DeInfoReg when training on the IMDB dataset with a batch size of 256 using LSTM as the backbone network. The results show that the four methods exhibit comparable GPU memory footprints. Specifically, BP uses 18088 MiB, AL consumes 18116 MiB, SCPL requires 18184 MiB, and DeInfoReg utilizes 18104 MiB. These differences are minimal, indicating that despite architectural modifications and the integration of decoupled supervised contrastive learning in DeInfoReg, its memory requirements remain in the same range as traditional backpropagation and other gradient decoupling approaches. This demonstrates that our method achieves improved performance and robustness without incurring significant additional GPU memory costs.

\subsection{The Impact of Local Loss Functions} \label{sec:loss-func-exp}

\begin{table}[tb]
    \centering
    \caption{Test accuracy of DeInfoReg with different local loss function components}
    \label{tab:local-loss-acc}
    \begin{tabular}{c||cc}
    \toprule
    Loss Combination & VGG (CIFAR-100) & LSTM (IMDB) \\ 
    \midrule
    Variance Only            & $73.06 \pm 0.38$ & $83.42 \pm 0.17$ \\
    Invariance Only          & $74.02 \pm 0.22$ & $89.90 \pm 0.14$ \\
    Covariance Only          & $72.89 \pm 0.26$ & $83.58 \pm 0.14$ \\
    Variance + Invariance    & $74.10 \pm 0.30$ & $90.10 \pm 0.20$ \\
    Variance + Covariance    & $73.20 \pm 0.28$ & $84.00 \pm 0.20$ \\
    Invariance + Covariance  & $74.00 \pm 0.25$ & $90.00 \pm 0.25$ \\
    All & $74.27 \pm 0.29$ & $90.36 \pm 0.27$ \\
    \bottomrule
    \end{tabular}
\end{table}

This section reports on an ablation study on the local loss functions.

We evaluate the impact of different local loss functions, variance, invariance, and covariance, of the DeInfoReg model based on the final accuracy. We test VGG on CIFAR-100 and LSTM on IMDB datasets.

Table \ref{tab:local-loss-acc} shows the result. When using only a single local loss, invariance is the most important. This makes sense as the invariance loss aims to minimize the distance between the predicted embedding and the target label distribution, which is probably the most important clue for each module. However, combining all three local losses yields the best results. For VGG on CIFAR-100, the accuracy improves from $74.02 \%$ to $74.27 \%$. For LSTM on IMDB, the accuracy improves from $89.90 \%$ to $90.36 \%$. These results highlight the hybrid effect of combining all three local loss functions.

\subsection{The Impact of the Relative Importance of Cross-entropy and Local Losses} \label{sec:alpha-value-exp}

This section aims to investigate the impact of the relative importance between the cross-entropy loss ($L^{CE}$) and the local loss ($L^{LL}$) in Equation~\ref{eq:loss_total}. This relative importance is governed by the hyperparameter $\alpha$. To assess the effect of different $\alpha$ values, we conducted a series of experiments using the DeInfoReg model with a VGG network on the CIFAR-100 dataset. The values of $\alpha$ are set to $1 \times 10^{-0}$, $1 \times 10^{-1}$, $1 \times 10^{-2}$, $1 \times 10^{-3}$, $1 \times 10^{-4}$, and $1 \times 10^{-5}$.

The experimental results, summarized in Table \ref{tab:alpha_impact}, demonstrate that varying the value of $\alpha$ can influence the accuracy of the model. The peak performance is observed when $\alpha$ is around $1 \times 10^{-2}$ and $1\times 10^{-3}$, which yields the optimal test accuracies. Increasing or decreasing the value of $\alpha$ is detrimental.

\begin{table}[tb]
\centering
\caption{Best test accuracy of DeInfoReg (VGG on CIFAR-100) with varying $\alpha$ values.}
\label{tab:alpha_impact}
\begin{tabular}{cc}
\toprule
$\alpha$ & Best Test Accuracy (\%) \\
\midrule
$1 \times 10^{-0}$ & 67.93 \\
$1 \times 10^{-1}$ & 70.43 \\
$1 \times 10^{-2}$ & 73.28 \\
$1 \times 10^{-3}$ & 72.66 \\
$1 \times 10^{-4}$ & 67.10 \\
$1 \times 10^{-5}$ & 47.91 \\
\bottomrule
\end{tabular}
\end{table}

The observed trend in performance with varying $\alpha$ values highlights a crucial trade-off in the DeInfoReg framework.
When $\alpha$ is too large (e.g., $1 \times 10^{-0}$), the cross-entropy loss ($L^{CE}$) term dominates the total loss for each module. This excessive emphasis on the final classification task perhaps forces intermediate modules to prematurely specialize their features for classification, potentially neglecting the development of rich, diverse, and generalizable representations that the local loss ($L^{LL}$) terms are designed to foster. Such premature specialization might lead to suboptimal feature hierarchies in the earlier layers, thereby limiting the overall capacity and performance of the decoupled learning system. Conversely, when $\alpha$ is too small (e.g., $1 \times 10^{-5}$), the contribution of the cross-entropy loss becomes negligible, and the training of each module is almost exclusively driven by the local loss $L^{LL}$. Although $L^{LL}$ promotes feature diversity, insufficient guidance from the global classification objective can be detrimental.

\section{Discussion} \label{sec:disc}

This study introduces the Decoupled Supervised Learning with Information Regularization (DeInfoReg) model architecture. DeInfoReg integrates the concept of contrastive learning and regularization losses to decouple the gradient flow of a neural network into multiple shorter ones, making optimization more efficient. Our experiments on open datasets for both visual and natural language tasks confirm that DeInfoReg consistently exhibits stable and outstanding performance, regardless of batch size, module depth, or noise robustness. Moreover, our results indicate that DeInfoReg effectively mitigates the vanishing gradient problem, a critical challenge in training deep networks.

In addition to its optimization benefits, the decoupled design of DeInfoReg naturally facilitates model parallelism. By partitioning the network into independent components that can perform forward and backward passes concurrently, DeInfoReg enables multi-GPU training that significantly improves training throughput. This ability to leverage parallel training mechanisms is particularly advantageous in large-scale settings, where computational resources can be fully utilized to reduce overall training time.

In standard BP, all parameters are optimized to ensure that the prediction $\hat{y}$ closely matches the ground truth $y$ In contrast, DeInfoReg -- and other models based on the decoupled paradigm -- employ gradient truncation, meaning that the parameter updates may not necessarily align directly with the ultimate objective of achieving $\hat{y} \approx y$. As a result, DeInfoReg may sometimes perform worse than BP on specific datasets, e,g., CIFAR-10. However, we emphasize that in other datasets, DeInfoReg typically stands out as the best-performing model among decoupled paradigm approaches and sometimes even surpasses BP. Furthermore, many default settings in neural networks, such as the choice of activation functions and learning rate schedulers, are primarily designed for standard BP and may not be optimal for decoupled models. Future research may explore better configurations explicitly tailored for decoupled approaches.

Other promising directions deserve further exploration as future work. First, integrating DeInfoReg with residual connections could be a valuable enhancement, as it may further improve gradient flows and mitigate issues related to vanishing or exploding gradients, thereby boosting the model's overall stability and performance in very deep networks. Second, the decoupled architecture makes it possible to implement asynchronous training strategies across multiple GPUs and further refine model parallelism, which could lead to even greater improvements in training efficiency. Third, we aim to develop an API similar to scikit-learn for the DeInfoReg model, providing users with greater flexibility in building and extending modules, simplifying customization and deployment, and facilitating integration into various workflows. Finally, testing DeInfoReg on larger datasets will be essential to validate its effectiveness in big data scenarios.

\section*{Declaration of generative AI and AI-assisted technologies in the writing process}

During the preparation of this work, the authors used ChatGPT and Gemini to improve language and readability. The authors reviewed and edited the content as needed and take full responsibility for the content of the publication.

\section*{Acknowledgement}
We acknowledge support from the National Science and Technology Council of Taiwan under grant number 113-2221-E-008-100-MY3. We thank to National Center for High-performance Computing (NCHC) of National Applied Research Laboratories (NARLabs) in Taiwan for providing computational and storage resources.

\bibliographystyle{plainnat}

\end{document}